\documentclass[sigconf, dvipsnames]{acmart}
\AtBeginDocument{%
  }

\usepackage[T1]{fontenc}
\usepackage{graphicx}
\usepackage{hyperref}
\usepackage{xcolor}
\usepackage[most]{tcolorbox}
\begin{document}

\title{How Effective is GPT-4 Turbo in Generating School-Level Questions from Textbooks Based on Bloom's Revised Taxonomy?}

\author{Subhankar Maity}
\orcid{0009-0001-1358-9534}
%\email{webmaster@marysville-ohio.com}
\affiliation{%
 \institution{IIT Kharagpur}
  %\streetaddress{Kharagpur}
   %\city{Kharagpur}
      \state{West Bengal}
  \country{India}
}
\email{subhankar.ai@kgpian.iitkgp.ac.in}
\authornote{Corresponding Author}

\author{Aniket Deroy}
\orcid{0000-0001-7190-5040}
\affiliation{
 \institution{IIT Kharagpur}
  %\streetaddress{1 Th{\o}rv{\"a}ld Circle}
  %\city{Hekla}
 %\city{Kharagpur}
\state{West Bengal}
\country{India}
}
\email{roydanik18@kgpian.iitkgp.ac.in}

\author{Sudeshna Sarkar}
\orcid{0000-0003-3439-4282}
\affiliation{%
 \institution{IIT Kharagpur}
  %\city{Rocquencourt}
 %\city{Kharagpur}
 \state{West Bengal}
 \country{India}
 }
\email{sudeshna@cse.iitkgp.ac.in}

%\renewcommand{\shortauthors}{Maity et al.}

\begin{abstract}
  We evaluate the effectiveness of GPT-4 Turbo in generating educational questions from NCERT textbooks in zero-shot mode. Our study highlights GPT-4 Turbo's ability to generate questions that require higher-order thinking skills, especially at the "understanding" level according to Bloom’s Revised Taxonomy. While we find a notable consistency between questions generated by GPT-4 Turbo and those assessed by humans in terms of complexity, there are occasional differences. Our evaluation also uncovers variations in how humans and machines evaluate question quality, with a trend inversely related to Bloom’s Revised Taxonomy levels. These findings suggest that while GPT-4 Turbo is a promising tool for educational question generation, its efficacy varies across different cognitive levels, indicating a need for further refinement to fully meet educational standards.
\end{abstract}


%\ccsdesc[500]{Computing methodologies~Natural language generation}
%\ccsdesc[300]{Computing methodologies~Language resources}
\keywords{Automated Question Generation (AQG), Large Language Models (LLMs), Bloom's Revised Taxonomy, GPT }

\maketitle

\section{Introduction}
In the realm of education, crafting high-quality questions stands as a pivotal task for educators striving to cultivate profound understanding and critical thinking among their students \cite{r46}. Questions serve as the cornerstone of learning assessments, acting as gateways to gauge understanding, recalling, application, analysis, evaluation, and creation, essentially mirroring the diverse levels of cognitive skills outlined in Bloom’s Revised Taxonomy \cite{r23}. This taxonomy provides educators with a structured framework to categorize and assess the depth and complexity of learning objectives, enabling them to tailor instructional strategies to meet the varied needs of learners. However, the manual process of question design often proves arduous and time-consuming for educators, demanding meticulous attention to detail and a deep understanding of pedagogical principles \cite{r46}. 

In recent years, the emergence of automated question generation (AQG), propelled by advancements in Artificial Intelligence (AI) and Natural Language Processing (NLP), has promised to revolutionize this fundamental aspect of education \cite{r53}. Large Language Models (LLMs), such as GPT-4 Turbo, have showcased remarkable capabilities in generating human-like text and responses across various domains \cite{r5}. Leveraging these LLM-driven technologies, educators can potentially streamline the question creation process, freeing up valuable time to focus on instructional delivery and student engagement.

In this study, we embark on a journey to explore the efficacy of an LLM-driven approach for generating and evaluating questions in the context of school-level education. To achieve this, we used zero-shot prompting with GPT-4 Turbo to generate questions from selected chapters of National Council of Educational Research
and Training (NCERT)\footnote{\url{https://en.wikipedia.org/wiki/National_Council_of_Educational_Research_and_Training}}  textbooks, evaluating the questions' alignment with Bloom's Revised Taxonomy. Focusing on subjects such as History, Geography, Economics, Environmental Studies, and Science, from the $6^{th}$ to the $12^{th}$ standard, we employ a multifaceted methodology to achieve our research objectives.

Our approach encompasses three key components. 

\begin{enumerate}
    \item We utilize the capabilities of GPT-4 Turbo in zero-shot mode to stimulate the creation of educational questions that correspond to Bloom’s Revised Taxonomy, with a focus on ensuring contextual appropriateness within specific fields of study. 

    \item We utilize advanced NLP methods to evaluate the quality of the questions produced, examining their compatibility with Bloom’s Revised Taxonomy and their agreement to Item Writing Flaw (IWF) criteria \cite{r47, r49}—a widely recognized set of guidelines in educational evaluation.

    \item To narrow the gap between automated evaluations and human expectations, a subset of the generated questions undergoes rigorous scrutiny by school-level teachers with expertise and pedagogical insights.

\end{enumerate}

This work aims to investigate these two research questions (RQs).

\begin{itemize}
    
    \item \textbf{RQ1}: What is the ability of GPT-4 Turbo to generate questions that align with the various levels of Bloom's Revised Taxonomy, as evaluated separately by a machine learning model and an educator?

    \item \textbf{RQ2}: How do human experts (i.e., Human-Validation) and machine validation using the IWF criteria (i.e., Machine-Validation) compare in assessing the quality of questions generated by GPT-4 Turbo, and what level of agreement exists between these two validation approaches?
    
\end{itemize}

These RQs guide our investigation into the effectiveness and dependability of employing advanced LLMs to generate and authenticate high-quality questions in educational evaluations.

\section{Related Work}
Recent research in question generation (QG) has focused on leveraging transformer-based Large Language Models (LLMs). These LLMs, deeply rooted in machine learning, undergo extensive training on large datasets to enhance their ability to generate text \cite{r1, r55}. The rationale behind adopting this approach in QG exploration primarily stems from its significant advancements in performance compared to previous rule-based and alternative systems \cite{r2, r3, r4}. During the training of transformer-based LLMs, the typical objective is to predict the next token, enabling these models to anticipate a likely continuation of an initial input text. Recent progress has witnessed the integration of reinforcement learning into the training methodologies of LLMs, such as GPT-4 Turbo \cite{r5}, which is used in the experiments discussed in this article. Fine-tuning through reinforcement learning, guided by human feedback, empowers these LLMs to surpass their predecessors \cite{r5}.

Aligned with the conventional training objective of LLMs, which involves predicting the next token, the evolving approach in QG entails providing a textual input, known as a prompt, to an LLM to generate completions \cite{r3, r54}. Crafting such prompts to elicit desired output can be challenging, leading to the emergence of a new research area known as prompt engineering. One prevalent method in prompt engineering is to prepend a string to the context provided to an LLM for generation, termed a prefix-style prompt \cite{r8}. For instance, consider a life science instructor aiming to generate questions related to photosynthesis. One straightforward approach they might take is to prompt an LLM with the following input: “\textit{Generate a question about photosynthesis}”. To refine the precision of the generated questions, the instructor could offer additional context. For example, instead of a general prompt, they could formulate a prompt containing a textbook excerpt focusing on a specific aspect of photosynthesis, such as: “\textit{Given the context <context>, generate a Question}” \cite{r9}. To exert more influence over the generation process, the teacher’s input could incorporate a control component—a keyword that guides the generation \cite{r3}. For instance, they might prompt an LLM with the following prompt: “\textit{Generate multiple-choice questions based on the given context <context> along with the correct answer and three distractors}” \cite{r10}.

Another facet of prompt engineering involves integrating examples that illustrate the desired output format and style directly into the prompt. This method, commonly known as few-shot learning, typically consists of an instruction, multiple examples, and the assigned task. These examples aim to acquaint LLMs with new contexts without requiring further training or fine-tuning \cite{r8}. \cite{r11} experimented with various prompting strategies to enhance educational question generation. Their study revealed that utilizing shorter input contexts and employing few-shot learning resulted in higher-quality candidate questions. They specifically focused on the GPT-2 model and utilized the SQuAD dataset \cite{r33} and OpenStax textbooks \cite{r41} for their experiments. \cite{r38} explored AQG using the OpenStax College Algebra textbook \cite{r40} as a reference. They investigated ChatGPT, particularly employing a zero-shot prompt to facilitate the generation process. The zero-shot prompt used in their experiment was: “\textit{Please generate 20 exercise questions based on this textbook chapter}”. 

Previous approaches have not conducted post-generational validation of Bloom's Revised Taxonomy, thus lacking definitive evidence regarding the effectiveness of their methods in generating questions that align accurately with Bloom’s Revised Taxonomy levels. Furthermore, there is a lack of exploration into the use of GPT-4 Turbo for generating educational questions according to Bloom's Revised Taxonomy, and no evaluation of questions using IWF criteria has been conducted in the aforementioned studies.

\section{Dataset}

In this study, we utilized the EduProbe dataset introduced by \cite{r9}, which is derived from the NCERT textbooks. These textbooks cover various subjects, including History, Geography, Economics, Environmental Studies, and Science, for grade levels ranging from $6^{th}$ to $12^{th}$ standard. For our experiments, we extracted 1,005 <Context, Question> pairs from this dataset.

\section{Methodology}

Bloom’s revised taxonomy \cite{r23} aids educators in creating educational questions tailored to specific learning objectives \cite{r34, r35, r37}. In this
method, we focus on generating questions aligned with Bloom’s revised taxonomy, a hierarchical framework that categorizes cognitive skills into remembering, understanding, applying, analyzing, evaluating, and creating. We utilize a zero-shot learning approach, designing prompts to elicit questions that correspond to each cognitive level of Bloom’s revised taxonomy. GPT-4 Turbo is then prompted with the zero-shot prompt (as illustrated in Figure \ref{fig1}), and the generated questions are evaluated for their alignment with Bloom’s revised taxonomy. Instead of generating one question and its taxonomic level sequentially, all six questions relevant to a specific context are generated simultaneously, as shown in Figure \ref{fig1}.

\begin{figure}[h]
    \centering
     \begin{tcolorbox}
     [enhanced, fit to height=3cm, 
     colback=blue!5!white, colframe=blue!!white, drop fuzzy shadow]
     %[colback=red!5!white,colframe=red!75!black]

      {\fontfamily{qcr}\selectfont

      Generate question at each level of Bloom’s revised taxonomy.
      \\
    \\
Context: \{context\}

Remembering:

Understanding:

Applying:

Analyzing:

Evaluating:

Creating:
}

\end{tcolorbox}

 \caption{Prompt template for generating questions following Bloom’s revised taxonomy in the zero-shot setting.}
    \label{fig1}
\end{figure}

\section{Evaluation}

\subsection{Evaluation of Question Quality}

We employed various NLP-based methods to automatically assess the educational quality of the generated questions by detecting common item writing flaws found in generated questions. The detection system reproduces the IWF detector described in \cite{r47, r49}. 

\subsection{Strategy for Evaluating Bloom's Revised Taxonomy}

We implemented the machine learning (ML) model described in \cite{r52} to classify the questions according to Bloom's Revised Taxonomy, ensuring that GPT-4 Turbo accurately generates questions at the correct Bloom's Revised Taxonomy level. The training dataset \cite{r52} encompasses questions across various subjects, such as natural sciences (e.g., Biology and Physics) and social sciences (e.g., Economics, History, and Sociology). It is designed to reflect the content commonly found in high school and college courses \cite{r52}. We divided the dataset into 85\% for training and 15\% for testing. The model achieved 87.89\% accuracy and 85.59\% weighted F1 score on the test set.

%To accurately categorize the questions' Bloom Revised Taxonomy-level with GPT-4 Turbo, we reproduced a machine learning (ML) model to classify each question’s Bloom’s Revised Taxonomy level, as discussed in \cite{r52}. 

\subsection{Human Evaluation}

We recruited five experts for five respective subjects with more than 20 years of teaching acquaintance to review the generated questions and verify the automated evaluation system. Each expert reviewed 60 randomly selected questions from a pool of 150, along with the contexts used to create them. They were asked to determine if they would utilize the generated questions during class (i.e., yes/no), determine Bloom's revised taxonomy level of every question, and evaluate whether the questions were relevant to their respective contexts (i.e., yes/no) \cite{r56}.

\section{Results}

\noindent \textbf{RQ1: What is the ability of GPT-4 Turbo to generate questions that align with the various levels of Bloom's Revised Taxonomy, as evaluated separately by a machine learning model and an educator?}

\begin{figure*}[h]
  \centering
  \includegraphics[width=\linewidth]{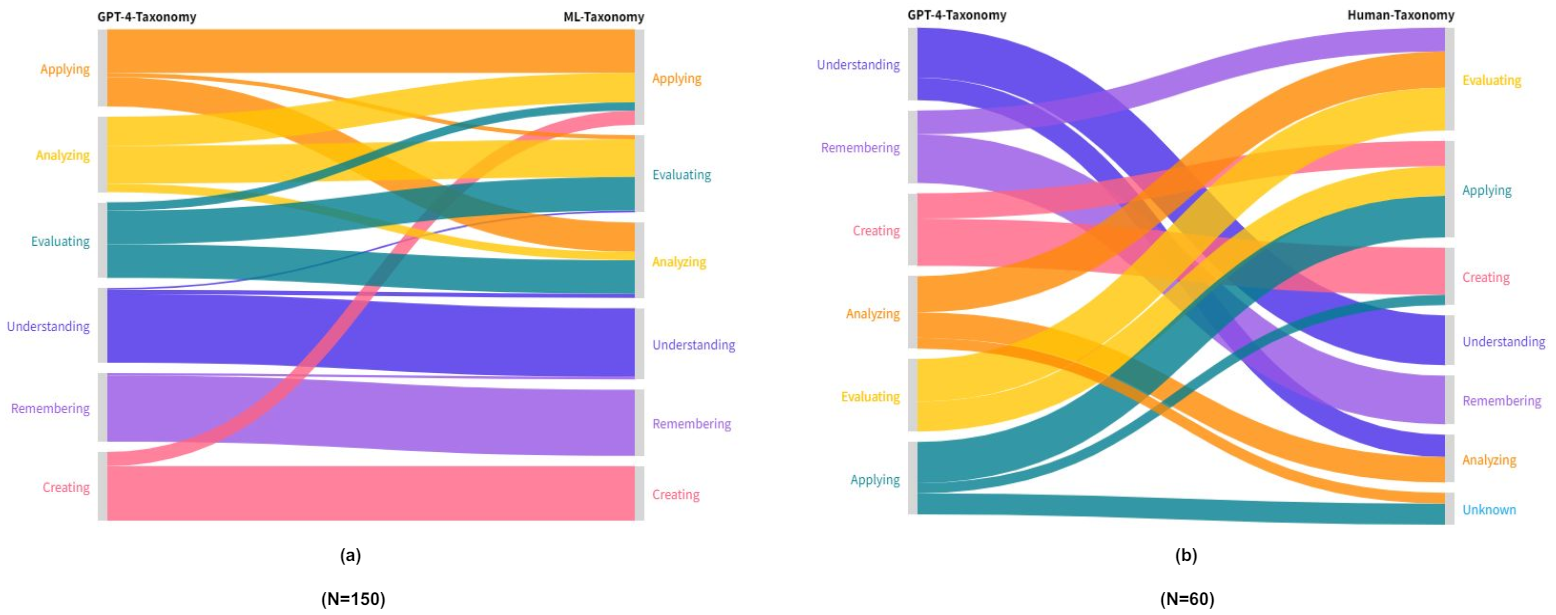}
  \caption{The level of alignment in Bloom’s Revised Taxonomy (a) between the GPT-4-Taxonomy and ML-Taxonomy (w/ 150 samples) and (b) between the GPT-4-Taxonomy and Human-Taxonomy (w/ 60 samples).} 
  \label{s1}
\end{figure*}

In this section, we analyze Bloom’s revised taxonomy levels of the questions produced by GPT-4 Turbo. We compare the levels identified for generation by GPT-4 Turbo (referred to as GPT-4-Taxonomy) with those projected by ML models (referred to as ML-Taxonomy) and the levels allocated by a school teacher (referred to as Human-Taxonomy). Figure \ref{s1} illustrates a Sankey diagram \cite{r48}, which represents the degree of correspondence between the GPT-4-Taxonomy, the ML-Taxonomy, and the Human-Taxonomy. Subplot (a) specifically shows the alignment between the GPT-4-Taxonomy and the ML-Taxonomy. Notably, 82\% of questions classified as "Understanding" by GPT-4 Turbo are also categorized at the same level by the ML model, while 79\% of "Remembering" questions and 65\% of "Creating" questions align similarly with the ML model's classifications. However, questions categorized as "Applying" are primarily split between "Applying" (52\%) and "Analyzing" (35\%) by the ML model, and "Analyzing" level questions are primarily classified as "Applying" (35\%) and "Evaluating" (45\%). For "Evaluating" questions, 40\% are categorized at the same level, while 40\% are categorized as "Analyzing" by the ML model. The ML model (i.e., ML-Taxonomy) predominantly assigns questions to the "Understanding" and "Remembering" levels, possibly due to the relatively unbalanced training set, where a majority of questions (37\%) are at these levels.

The alignment between GPT-4-Taxonomy and Human-Taxonomy is depicted in subplot (b) (Figure \ref{s1}). There is a notable level of agreement, especially for the levels of understanding, remembering, and creating. Specifically, 69\%, 67\%, and 65\% of the questions classified under these levels by GPT-4 Turbo were similarly categorized by the teacher. For example, about 50\% of the questions marked as "Analyzing" by GPT-4 Turbo were rated as "Evaluating" by the teacher. Additionally, it is worth mentioning that the human rater was unable to assign specific levels to a few questions initially categorized as "Applying" and "Analyzing" by GPT-4 Turbo (i.e., Unknown).

\noindent \textbf{RQ2: How do human experts (i.e., Human-Validation) and machine validation using the IWF criteria (i.e., Machine-Validation) compare in assessing the quality of questions generated by GPT-4 Turbo, and what level of agreement exists between these two validation approaches?}

\begin{figure*}[h]
  \centering
  \includegraphics[width=\linewidth]{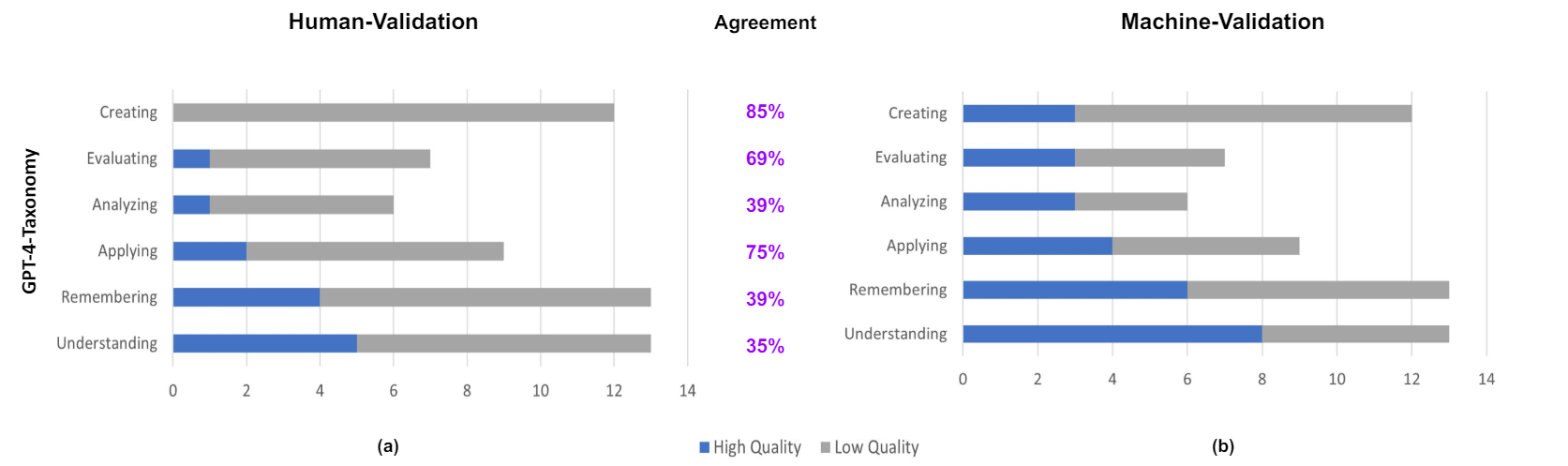}
  \caption{Quality evaluation outcomes for a sample of 60 questions generated by GPT-4-Turbo (i.e., GPT-4-Taxonomy) were evaluated by (a) a human teacher (i.e., Human-Validation) and (b) an ML model according to the IWF criteria (i.e., Machine-Validation), as well as the agreement between the two validation approaches.} 
  \label{rq2}
\end{figure*}

Figure \ref{rq2} illustrates the results of quality evaluation for a subset of 60 questions generated by GPT-4 Turbo. These questions were assessed by both a human teacher (shown in subplot (a)) and a machine (shown in subplot (b)) using the IWF criteria. The teacher's evaluation is binary, determining whether a question is of high quality and appropriate for practical use. In contrast, the machine's assessment considers a question to be of high standard if it meets at least nine IWF criteria \cite{r49}; otherwise, it is regarded as low quality. Based on human assessment, only 13 out of the 60 questions are judged to be of high-quality, primarily based on "understanding" and "remembering" cognitive levels (according to GPT-4-Taxonomy). However, the IWF criteria classified 27 questions (45\%) as high quality, with a substantial portion also from the "understanding" and "remembering" cognitive levels. At the evaluation level, both human and machine evaluations find it difficult to identify high-quality questions. The agreement between human and machine assessments varied considerably, ranging from 35\% for questions at the analyzing level to 85\% for those at the evaluating level.

\section{Analysis}

The investigation of RQ1 provides insight into the correlation between the questions generated by GPT-4 Turbo and the levels of Bloom's revised taxonomy, as evaluated separately by an ML model (i.e., ML-Taxonomy) and a school teacher (i.e., Human-Taxonomy). The results illuminate both the advantages and the limitations of employing AQG in educational evaluations.

The evaluation in RQ1 demonstrates the capability of GPT-4 Turbo to create questions that adhere to the desired Bloom’s Revised Taxonomy levels. In particular, a significant alignment is observed, particularly at the "understanding" level, indicating GPT-4 Turbo’s proficiency in generating questions that require higher-order cognitive skills. Nevertheless, certain discrepancies, notably between closely linked levels such as "Analyzing" and "Evaluating," underscore the challenge in distinguishing between these cognitive levels. These inconsistencies emphasize the necessity for ongoing improvement in GPT-4 Turbo's question generation capabilities.

A comparison of GPT-4-Taxonomy and Human-Taxonomy demonstrates considerable agreement, especially regarding the levels of "Understanding," "Remembering," and "Creating." This correspondence indicates that GPT-4 Turbo is generally adept at producing questions that match human views on its complexity. However, occasional discrepancies exist, such as questions categorized as "Analyzing" by GPT-4 Turbo being classified as "Evaluating" by the teacher, highlighting areas where enhancements in distinguishing between higher-order cognitive levels could be beneficial. Additionally, some questions were challenging for teachers to categorize definitively, suggesting ambiguity within a subset of the generated questions.

The investigation of RQ2 revolves around the evaluation of the standard of the questions produced by GPT-4 Turbo using two validation methods: \textit{Human-Validation} and \textit{Machine-Validation} employing IWF criteria. Furthermore, this examination explores the correlation between these validation approaches. Findings indicate that a small fraction of the generated questions, specifically 13 out of 60, meet high standards according to Human-Validation. These questions primarily pertain to the "Understanding" and "Remembering" levels, consistent with Bloom’s Revised Taxonomy as suggested by GPT-4 Turbo. Conversely, machine evaluation using IWF criteria identified a larger percentage—27 questions or 45\%—as high quality, with prominence in questions from cognitive levels of "Understanding" and "Remembering".

An interesting finding is the reverse relationship observed between Bloom’s Revised Taxonomy levels and question quality. GPT-4 Turbo faces growing hurdles in producing high-quality questions. For example, at the "Evaluating" level, both human teachers and machines struggled more to pinpoint questions of exceptional quality. This suggests that higher levels of cognitive complexity, like "Analyzing" and "Evaluating," present more formidable obstacles for AQG.

\section{Conclusion and Future Work}

The outcomes of RQ1 show that GPT-4 Turbo can generate questions that correspond with the levels of Bloom's revised taxonomy. They emphasize the need to enhance GPT-4 Turbo's question generation abilities, address nuances in cognitive level distinctions, and ensure alignment with human standards in educational settings. These results lay the groundwork for further exploration of LLM-driven AQG for educational assessment purposes. Additionally, findings from RQ2 reveal the complexities of AQG and quality assessment, highlighting the differences between human teacher and machine evaluations. This emphasizes the importance of a holistic approach that considers both perspectives. The observed inverse correlation between cognitive complexity and question quality highlights the difficulty in crafting superior-quality questions at more advanced levels of taxonomy. These insights indicate ways to improve automated systems so they better meet human standards and expectations in the creation of educational content.

%These insights suggest avenues for enhancing automated systems to align more closely with human standards and expectations in the creation of educational material.

We intend to employ few-shot learning with eight-shot, a methodology introduced by \cite{r11}, to improve the quality of the generated questions. We aim to refine the automated validation of question quality by constructing an ML model utilizing IWF features, aiming at a closer approximation to human verification. Furthermore, we plan to further refine Bloom’s revised taxonomy model by leveraging GPT-4-Turbo-generated questions, annotated with the assistance of Mechanical Turk workers. The evaluation of questions will be performed using our existing assessment framework, in conjunction with linguistic quality metrics such as perplexity and diversity scores, and consideration of the relevance of the questions within specific contexts.
\bibliographystyle{ACM-Reference-Format}
\bibliography{sample-base}


\begin{thebibliography}{26}

%%% the Web address.

\ifx \showCODEN    \undefined \def \showCODEN     #1{\unskip}     \fi
\ifx \showDOI      \undefined \def \showDOI       #1{#1}\fi
\ifx \showISBNx    \undefined \def \showISBNx     #1{\unskip}     \fi
\ifx \showISBNxiii \undefined \def \showISBNxiii  #1{\unskip}     \fi
\ifx \showISSN     \undefined \def \showISSN      #1{\unskip}     \fi
\ifx \showLCCN     \undefined \def \showLCCN      #1{\unskip}     \fi
\ifx \shownote     \undefined \def \shownote      #1{#1}          \fi
\ifx \showarticletitle \undefined \def \showarticletitle #1{#1}   \fi
\ifx \showURL      \undefined \def \showURL       {\relax}        \fi
% invisible to TeX
\providecommand\bibfield[2]{#2}
\providecommand\bibinfo[2]{#2}
\providecommand\natexlab[1]{#1}
\providecommand\showeprint[2][]{arXiv:#2}

\bibitem[Abramson(2021)]%
        {r40}
\bibfield{author}{\bibinfo{person}{Jay Abramson}.} \bibinfo{year}{2021}\natexlab{}.
\newblock \bibinfo{booktitle}{\emph{College Algebra 2e with Corequisite Support}}.
\newblock \bibinfo{publisher}{OpenStax}.
\newblock


\bibitem[Al~Faraby et~al\mbox{.}(2023)]%
        {r37}
\bibfield{author}{\bibinfo{person}{Said Al~Faraby}, \bibinfo{person}{Adiwijaya Adiwijaya}, {and} \bibinfo{person}{Ade Romadhony}.} \bibinfo{year}{2023}\natexlab{}.
\newblock \showarticletitle{Review on neural question generation for education purposes}.
\newblock \bibinfo{journal}{\emph{International Journal of Artificial Intelligence in Education}} (\bibinfo{year}{2023}), \bibinfo{pages}{1--38}.
\newblock


\bibitem[Bhandari et~al\mbox{.}(2023)]%
        {r38}
\bibfield{author}{\bibinfo{person}{Shreya Bhandari}, \bibinfo{person}{Yunting Liu}, {and} \bibinfo{person}{Zachary~A Pardos}.} \bibinfo{year}{2023}\natexlab{}.
\newblock \showarticletitle{Evaluating ChatGPT-generated Textbook Questions using IRT}. In \bibinfo{booktitle}{\emph{Proceedings of the Generative AI for Education Workshop (GAIED) at the Thirty-seventh Conference on Neural Information Processing Systems (NeurIPS). New Orleans, LA}}.
\newblock


\bibitem[Bloom(2010)]%
        {r23}
\bibfield{author}{\bibinfo{person}{Benjamin~Samuel Bloom}.} \bibinfo{year}{2010}\natexlab{}.
\newblock \bibinfo{booktitle}{\emph{A taxonomy for learning, teaching, and assessing: A revision of Bloom's taxonomy of educational objectives}}.
\newblock \bibinfo{publisher}{Longman}.
\newblock


\bibitem[Edward J.~Palmer and Russell(2010)]%
        {r49}
\bibfield{author}{\bibinfo{person}{Peter G.~Devitt Edward J.~Palmer, Paul~Duggan} {and} \bibinfo{person}{Rohan Russell}.} \bibinfo{year}{2010}\natexlab{}.
\newblock \showarticletitle{The modified essay question: Its exit from the exit examination?}
\newblock \bibinfo{journal}{\emph{Medical Teacher}} \bibinfo{volume}{32}, \bibinfo{number}{7} (\bibinfo{year}{2010}), \bibinfo{pages}{e300--e307}.
\newblock
\urldef\tempurl%
\url{https://doi.org/10.3109/0142159X.2010.488705}
\showDOI{\tempurl}
\showeprint{https://doi.org/10.3109/0142159X.2010.488705}
\newblock
\shownote{PMID: 20653373}.


\bibitem[Gong et~al\mbox{.}(2022)]%
        {r46}
\bibfield{author}{\bibinfo{person}{Huanli Gong}, \bibinfo{person}{Liangming Pan}, {and} \bibinfo{person}{Hengchang Hu}.} \bibinfo{year}{2022}\natexlab{}.
\newblock \showarticletitle{{KHANQ}: A Dataset for Generating Deep Questions in Education}. In \bibinfo{booktitle}{\emph{Proceedings of the 29th International Conference on Computational Linguistics}}, \bibfield{editor}{\bibinfo{person}{Nicoletta Calzolari}, \bibinfo{person}{Chu-Ren Huang}, \bibinfo{person}{Hansaem Kim}, \bibinfo{person}{James Pustejovsky}, \bibinfo{person}{Leo Wanner}, \bibinfo{person}{Key-Sun Choi}, \bibinfo{person}{Pum-Mo Ryu}, \bibinfo{person}{Hsin-Hsi Chen}, \bibinfo{person}{Lucia Donatelli}, \bibinfo{person}{Heng Ji}, \bibinfo{person}{Sadao Kurohashi}, \bibinfo{person}{Patrizia Paggio}, \bibinfo{person}{Nianwen Xue}, \bibinfo{person}{Seokhwan Kim}, \bibinfo{person}{Younggyun Hahm}, \bibinfo{person}{Zhong He}, \bibinfo{person}{Tony~Kyungil Lee}, \bibinfo{person}{Enrico Santus}, \bibinfo{person}{Francis Bond}, {and} \bibinfo{person}{Seung-Hoon Na}} (Eds.). \bibinfo{publisher}{International Committee on Computational Linguistics}, \bibinfo{address}{Gyeongju, Republic of Korea},
  \bibinfo{pages}{5925--5938}.
\newblock
\urldef\tempurl%
\url{https://aclanthology.org/2022.coling-1.518}
\showURL{%
\tempurl}


\bibitem[Gul et~al\mbox{.}(2020)]%
        {r35}
\bibfield{author}{\bibinfo{person}{Rani Gul}, \bibinfo{person}{Shazia Kanwal}, {and} \bibinfo{person}{Sadia~Suleman Khan}.} \bibinfo{year}{2020}\natexlab{}.
\newblock \showarticletitle{Preferences of the teachers in employing revised blooms taxonomy in their instructions}.
\newblock \bibinfo{journal}{\emph{sjesr}} \bibinfo{volume}{3}, \bibinfo{number}{2} (\bibinfo{year}{2020}), \bibinfo{pages}{258--266}.
\newblock


\bibitem[Hift(2014)]%
        {r47}
\bibfield{author}{\bibinfo{person}{Richard~J Hift}.} \bibinfo{year}{2014}\natexlab{}.
\newblock \showarticletitle{Should essays and other “open-ended”-type questions retain a place in written summative assessment in clinical medicine?}
\newblock \bibinfo{journal}{\emph{BMC Medical Education}}  \bibinfo{volume}{14} (\bibinfo{year}{2014}), \bibinfo{pages}{1--18}.
\newblock


\bibitem[Hwang et~al\mbox{.}(2023)]%
        {r56}
\bibfield{author}{\bibinfo{person}{Kevin Hwang}, \bibinfo{person}{Sai Challagundla}, \bibinfo{person}{Maryam Alomair}, \bibinfo{person}{Lujie~Karen Chen}, {and} \bibinfo{person}{Fow-Sen Choa}.} \bibinfo{year}{2023}\natexlab{}.
\newblock \showarticletitle{Towards AI-assisted multiple choice question generation and quality evaluation at scale: Aligning with Bloom’s Taxonomy}. In \bibinfo{booktitle}{\emph{Workshop on Generative AI for Education}}.
\newblock


\bibitem[Kurdi et~al\mbox{.}(2020)]%
        {r2}
\bibfield{author}{\bibinfo{person}{Ghader Kurdi}, \bibinfo{person}{Jared Leo}, \bibinfo{person}{Bijan Parsia}, \bibinfo{person}{Uli Sattler}, {and} \bibinfo{person}{Salam Al-Emari}.} \bibinfo{year}{2020}\natexlab{}.
\newblock \showarticletitle{A systematic review of automatic question generation for educational purposes}.
\newblock \bibinfo{journal}{\emph{International Journal of Artificial Intelligence in Education}}  \bibinfo{volume}{30} (\bibinfo{year}{2020}), \bibinfo{pages}{121--204}.
\newblock


\bibitem[Lee et~al\mbox{.}(2023)]%
        {r53}
\bibfield{author}{\bibinfo{person}{Unggi Lee}, \bibinfo{person}{Haewon Jung}, \bibinfo{person}{Younghoon Jeon}, \bibinfo{person}{Younghoon Sohn}, \bibinfo{person}{Wonhee Hwang}, \bibinfo{person}{Jewoong Moon}, {and} \bibinfo{person}{Hyeoncheol Kim}.} \bibinfo{year}{2023}\natexlab{}.
\newblock \showarticletitle{Few-shot is enough: exploring ChatGPT prompt engineering method for automatic question generation in english education}.
\newblock \bibinfo{journal}{\emph{Education and Information Technologies}} (\bibinfo{year}{2023}), \bibinfo{pages}{1--33}.
\newblock


\bibitem[Liu et~al\mbox{.}(2023)]%
        {r8}
\bibfield{author}{\bibinfo{person}{Pengfei Liu}, \bibinfo{person}{Weizhe Yuan}, \bibinfo{person}{Jinlan Fu}, \bibinfo{person}{Zhengbao Jiang}, \bibinfo{person}{Hiroaki Hayashi}, {and} \bibinfo{person}{Graham Neubig}.} \bibinfo{year}{2023}\natexlab{}.
\newblock \showarticletitle{Pre-train, Prompt, and Predict: A Systematic Survey of Prompting Methods in Natural Language Processing}.
\newblock \bibinfo{journal}{\emph{ACM Comput. Surv.}} \bibinfo{volume}{55}, \bibinfo{number}{9}, Article \bibinfo{articleno}{195} (\bibinfo{date}{jan} \bibinfo{year}{2023}), \bibinfo{numpages}{35}~pages.
\newblock
\showISSN{0360-0300}
\urldef\tempurl%
\url{https://doi.org/10.1145/3560815}
\showDOI{\tempurl}


\bibitem[Maity et~al\mbox{.}(2024a)]%
        {r54}
\bibfield{author}{\bibinfo{person}{Subhankar Maity}, \bibinfo{person}{Aniket Deroy}, {and} \bibinfo{person}{Sudeshna Sarkar}.} \bibinfo{year}{2024}\natexlab{a}.
\newblock \showarticletitle{Exploring the Capabilities of Prompted Large Language Models in Educational and Assessment Applications}.
\newblock \bibinfo{journal}{\emph{arXiv preprint arXiv:2405.11579}} (\bibinfo{year}{2024}).
\newblock


\bibitem[Maity et~al\mbox{.}(2024b)]%
        {r9}
\bibfield{author}{\bibinfo{person}{Subhankar Maity}, \bibinfo{person}{Aniket Deroy}, {and} \bibinfo{person}{Sudeshna Sarkar}.} \bibinfo{year}{2024}\natexlab{b}.
\newblock \showarticletitle{Harnessing the Power of Prompt-based Techniques for Generating School-Level Questions using Large Language Models}. In \bibinfo{booktitle}{\emph{Proceedings of the 15th Annual Meeting of the Forum for Information Retrieval Evaluation}} (, Panjim, India,) \emph{(\bibinfo{series}{FIRE '23})}. \bibinfo{publisher}{Association for Computing Machinery}, \bibinfo{address}{New York, NY, USA}, \bibinfo{pages}{30–39}.
\newblock
\showISBNx{9798400716324}
\urldef\tempurl%
\url{https://doi.org/10.1145/3632754.3632755}
\showDOI{\tempurl}


\bibitem[Maity et~al\mbox{.}(2024c)]%
        {r55}
\bibfield{author}{\bibinfo{person}{Subhankar Maity}, \bibinfo{person}{Aniket Deroy}, {and} \bibinfo{person}{Sudeshna Sarkar}.} \bibinfo{year}{2024}\natexlab{c}.
\newblock \showarticletitle{How Ready Are Generative Pre-trained Large Language Models for Explaining Bengali Grammatical Errors?}
\newblock \bibinfo{journal}{\emph{arXiv preprint arXiv:2406.00039}} (\bibinfo{year}{2024}).
\newblock


\bibitem[Maity et~al\mbox{.}(2024d)]%
        {r10}
\bibfield{author}{\bibinfo{person}{Subhankar Maity}, \bibinfo{person}{Aniket Deroy}, {and} \bibinfo{person}{Sudeshna Sarkar}.} \bibinfo{year}{2024}\natexlab{d}.
\newblock \showarticletitle{A Novel Multi-Stage Prompting Approach for Language Agnostic MCQ Generation Using GPT}. In \bibinfo{booktitle}{\emph{Advances in Information Retrieval}}, \bibfield{editor}{\bibinfo{person}{Nazli Goharian}, \bibinfo{person}{Nicola Tonellotto}, \bibinfo{person}{Yulan He}, \bibinfo{person}{Aldo Lipani}, \bibinfo{person}{Graham McDonald}, \bibinfo{person}{Craig Macdonald}, {and} \bibinfo{person}{Iadh Ounis}} (Eds.). \bibinfo{publisher}{Springer Nature Switzerland}, \bibinfo{address}{Cham}, \bibinfo{pages}{268--277}.
\newblock
\showISBNx{978-3-031-56063-7}


\bibitem[Mulla and Gharpure(2023)]%
        {r3}
\bibfield{author}{\bibinfo{person}{Nikahat Mulla} {and} \bibinfo{person}{Prachi Gharpure}.} \bibinfo{year}{2023}\natexlab{}.
\newblock \showarticletitle{Automatic question generation: a review of methodologies, datasets, evaluation metrics, and applications}.
\newblock \bibinfo{journal}{\emph{Progress in Artificial Intelligence}} \bibinfo{volume}{12}, \bibinfo{number}{1} (\bibinfo{year}{2023}), \bibinfo{pages}{1--32}.
\newblock


\bibitem[OpenAI(2023)]%
        {r5}
\bibfield{author}{\bibinfo{person}{OpenAI}.} \bibinfo{year}{2023}\natexlab{}.
\newblock \bibinfo{title}{GPT-4 Technical Report}.
\newblock
\newblock
\showeprint[arxiv]{2303.08774}~[cs.CL]


\bibitem[Otto et~al\mbox{.}(2022)]%
        {r48}
\bibfield{author}{\bibinfo{person}{Ethan Otto}, \bibinfo{person}{Eva Culakova}, \bibinfo{person}{Sixu Meng}, \bibinfo{person}{Zhihong Zhang}, \bibinfo{person}{Huiwen Xu}, \bibinfo{person}{Supriya Mohile}, {and} \bibinfo{person}{Marie~A. Flannery}.} \bibinfo{year}{2022}\natexlab{}.
\newblock \showarticletitle{Overview of Sankey flow diagrams: Focusing on symptom trajectories in older adults with advanced cancer}.
\newblock \bibinfo{journal}{\emph{Journal of Geriatric Oncology}} \bibinfo{volume}{13}, \bibinfo{number}{5} (\bibinfo{year}{2022}), \bibinfo{pages}{742--746}.
\newblock
\showISSN{1879-4068}
\urldef\tempurl%
\url{https://doi.org/10.1016/j.jgo.2021.12.017}
\showDOI{\tempurl}


\bibitem[Rajpurkar et~al\mbox{.}(2016)]%
        {r33}
\bibfield{author}{\bibinfo{person}{Pranav Rajpurkar}, \bibinfo{person}{Jian Zhang}, \bibinfo{person}{Konstantin Lopyrev}, {and} \bibinfo{person}{Percy Liang}.} \bibinfo{year}{2016}\natexlab{}.
\newblock \showarticletitle{{SQ}u{AD}: 100,000+ Questions for Machine Comprehension of Text}. In \bibinfo{booktitle}{\emph{Proceedings of the 2016 Conference on Empirical Methods in Natural Language Processing}}, \bibfield{editor}{\bibinfo{person}{Jian Su}, \bibinfo{person}{Kevin Duh}, {and} \bibinfo{person}{Xavier Carreras}} (Eds.). \bibinfo{publisher}{Association for Computational Linguistics}, \bibinfo{address}{Austin, Texas}, \bibinfo{pages}{2383--2392}.
\newblock
\urldef\tempurl%
\url{https://doi.org/10.18653/v1/D16-1264}
\showDOI{\tempurl}


\bibitem[Sideeg(2016)]%
        {r34}
\bibfield{author}{\bibinfo{person}{Abdunasir Sideeg}.} \bibinfo{year}{2016}\natexlab{}.
\newblock \showarticletitle{Bloom’s Taxonomy, Backward Design, and Vygotsky’s Zone of Proximal Development in crafting learning outcomes}.
\newblock \bibinfo{journal}{\emph{International Journal of Linguistics}} \bibinfo{volume}{8}, \bibinfo{number}{2} (\bibinfo{year}{2016}), \bibinfo{pages}{158--186}.
\newblock


\bibitem[Stafford and Flatley(2018)]%
        {r41}
\bibfield{author}{\bibinfo{person}{Daniel Stafford} {and} \bibinfo{person}{Robert Flatley}.} \bibinfo{year}{2018}\natexlab{}.
\newblock \showarticletitle{Openstax}.
\newblock \bibinfo{journal}{\emph{The Charleston Advisor}} \bibinfo{volume}{20}, \bibinfo{number}{1} (\bibinfo{year}{2018}), \bibinfo{pages}{48--51}.
\newblock


\bibitem[Steuer et~al\mbox{.}(2021)]%
        {r4}
\bibfield{author}{\bibinfo{person}{Tim Steuer}, \bibinfo{person}{Leonard Bongard}, \bibinfo{person}{Jan Uhlig}, {and} \bibinfo{person}{Gianluca Zimmer}.} \bibinfo{year}{2021}\natexlab{}.
\newblock \showarticletitle{On the linguistic and pedagogical quality of automatic question generation via neural machine translation}. In \bibinfo{booktitle}{\emph{Technology-Enhanced Learning for a Free, Safe, and Sustainable World: 16th European Conference on Technology Enhanced Learning, EC-TEL 2021, Bolzano, Italy, September 20-24, 2021, Proceedings 16}}. Springer, \bibinfo{pages}{289--294}.
\newblock


\bibitem[Wang et~al\mbox{.}(2021)]%
        {r52}
\bibfield{author}{\bibinfo{person}{Zichao Wang}, \bibinfo{person}{Kyle Manning}, \bibinfo{person}{Debshila~Basu Mallick}, {and} \bibinfo{person}{Richard~G. Baraniuk}.} \bibinfo{year}{2021}\natexlab{}.
\newblock \showarticletitle{Towards Blooms Taxonomy Classification Without Labels}. In \bibinfo{booktitle}{\emph{Artificial Intelligence in Education}}, \bibfield{editor}{\bibinfo{person}{Ido Roll}, \bibinfo{person}{Danielle McNamara}, \bibinfo{person}{Sergey Sosnovsky}, \bibinfo{person}{Rose Luckin}, {and} \bibinfo{person}{Vania Dimitrova}} (Eds.). \bibinfo{publisher}{Springer International Publishing}, \bibinfo{address}{Cham}, \bibinfo{pages}{433--445}.
\newblock
\showISBNx{978-3-030-78292-4}


\bibitem[Wang et~al\mbox{.}(2022)]%
        {r11}
\bibfield{author}{\bibinfo{person}{Zichao Wang}, \bibinfo{person}{Jakob Valdez}, \bibinfo{person}{Debshila Basu~Mallick}, {and} \bibinfo{person}{Richard~G. Baraniuk}.} \bibinfo{year}{2022}\natexlab{}.
\newblock \showarticletitle{Towards Human-Like Educational Question Generation with Large Language Models}. In \bibinfo{booktitle}{\emph{Artificial Intelligence in Education}}, \bibfield{editor}{\bibinfo{person}{Maria~Mercedes Rodrigo}, \bibinfo{person}{Noburu Matsuda}, \bibinfo{person}{Alexandra~I. Cristea}, {and} \bibinfo{person}{Vania Dimitrova}} (Eds.). \bibinfo{publisher}{Springer International Publishing}, \bibinfo{address}{Cham}, \bibinfo{pages}{153--166}.
\newblock
\showISBNx{978-3-031-11644-5}


\bibitem[Zhang et~al\mbox{.}(2023)]%
        {r1}
\bibfield{author}{\bibinfo{person}{Hanqing Zhang}, \bibinfo{person}{Haolin Song}, \bibinfo{person}{Shaoyu Li}, \bibinfo{person}{Ming Zhou}, {and} \bibinfo{person}{Dawei Song}.} \bibinfo{year}{2023}\natexlab{}.
\newblock \showarticletitle{A Survey of Controllable Text Generation Using Transformer-based Pre-trained Language Models}.
\newblock \bibinfo{journal}{\emph{ACM Comput. Surv.}} \bibinfo{volume}{56}, \bibinfo{number}{3}, Article \bibinfo{articleno}{64} (\bibinfo{date}{oct} \bibinfo{year}{2023}), \bibinfo{numpages}{37}~pages.
\newblock
\showISSN{0360-0300}
\urldef\tempurl%
\url{https://doi.org/10.1145/3617680}
\showDOI{\tempurl}


\end{thebibliography}

\end{document}